\documentclass[9pt,conference]{IEEEtran}
\pdfoutput=1
\IEEEoverridecommandlockouts
\usepackage{cite}
\usepackage{amsmath,amssymb,amsfonts}
\usepackage{algorithm2e}
\usepackage{graphicx}
\usepackage{textcomp}
\usepackage{xcolor}
\usepackage[hidelinks]{hyperref}
\usepackage{amssymb}
\usepackage{booktabs}
\usepackage{subcaption}
\usepackage{xspace}
\usepackage[a4paper, total={184mm,239mm}]{geometry}
\def\BibTeX{{\rm B\kern-.05em{\sc i\kern-.025em b}\kern-.08em
    T\kern-.1667em\lower.7ex\hbox{E}\kern-.125emX}}
    
\newcommand\mnade{TECO\xspace}
\newcommand*{\affaddr}[1]{#1} % No op here. Customize it for different styles.
\newcommand*{\affmark}[1][*]{\textsuperscript{#1}}

\SetKwComment{Comment}{/* }{ */}

\RestyleAlgo{ruled}

\usepackage{pdfrender}
\newcommand*{\boldcheckmark}{%
  \textpdfrender{
    TextRenderingMode=FillStroke,
    LineWidth=1pt, % half of the line width is outside the normal glyph
  }{\checkmark}%
}
    
\begin{document}

% \title{AdaptDL: Multi-Objective Model Adaption for Efficient On-Device Deep Learning

\title{Towards Efficient Convolutional Neural Network for Embedded Hardware via Multi-Dimensional Pruning\vspace{-10pt}
\thanks{\copyright~2023 IEEE. Personal use of this material is permitted. Permission from IEEE must be obtained for all other uses, in any current or future media, including reprinting/republishing this material for advertising or promotional purposes, creating new collective works, for resale or redistribution to servers or lists, or reuse of any copyrighted component of this work in other works. This is the author's accepted version of the article published in Proceedings of the 60th ACM/IEEE Design Automation Conference (DAC), pp. 1-6, 2023, DOI: 10.1109/DAC56929.2023.10247965.}
}

\author{
\\
Hao Kong\affmark[1,2], Di Liu\affmark[3], Xiangzhong Luo\affmark[1], Shuo Huai\affmark[1,2], Ravi Subramaniam\affmark[4], \\
 Christian Makaya\affmark[4], Qian Lin\affmark[4], and Weichen Liu\affmark[1]\affmark[*]
 
 \vspace{10pt}\\
\affaddr{\normalsize \affmark[1]School of Computer Science and Engineering, Nanyang Technological University, Singapore}\\
\affaddr{\normalsize \affmark[2]HP-NTU Digital Manufacturing Corporate Lab, Nanyang Technological University, Singapore}\\
\affaddr{\normalsize \affmark[3]Department of Computer Science, Norwegian University of Science and Technology, Trondheim, Norway.}\\
\affaddr{\normalsize \affmark[4]HP Inc., Palo Alto, California, USA}
}

% \and
% \and
% \and
% \and
% \and

\maketitle

\begin{abstract}

In this paper, we propose \mnade, a multi-dimensional pruning framework to collaboratively prune the three dimensions (depth, width, and resolution) of convolutional neural networks (CNNs) for better execution efficiency on embedded hardware. In \mnade, we first introduce a two-stage importance evaluation framework, which efficiently and comprehensively evaluates each pruning unit according to both the local importance inside each dimension and the global importance across different dimensions. Based on the evaluation framework, we present a heuristic pruning algorithm to progressively prune the three dimensions of CNNs towards the optimal trade-off between accuracy and efficiency.
Experiments on multiple benchmarks validate the advantages of \mnade over existing state-of-the-art (SOTA) approaches.
The code and pre-trained models are available anonymously at \textcolor{blue}{\url{https://github.com/ntuliuteam/Teco}}.
\end{abstract}

% \begin{IEEEkeywords}
% neural network pruning, edge computing, embedded systems, hardware/software co-design
% \end{IEEEkeywords}

\section{Introduction}

Over the past decade, the evolution of deep convolutional neural networks (CNNs) has been benefiting many deep learning applications \cite{he2016deep}. Recently, there is a growing demand to deploy advanced CNNs at the edge to address the concerns of network latency and data privacy \cite{liu2021bringing,shi2016edge}. However, modern CNNs are usually equipped with billions of operations. For example, the most popular CNN model, ResNet50 \cite{he2016deep}, has 4.1 billion Multiply-Accumulate operations (MACs), which are computationally prohibitive for embedded hardware \cite{kong2021edlab}.

To enable more edge deep learning applications, such as autopilot and smart cameras, to be benefited from the advances of CNNs, efforts have been made to compress CNNs for a better trade-off between execution efficiency and accuracy. Neural network pruning \cite{liu2021bringing, huang2018data, wang2018learning, molchanov2019importance, lin2020hrank, lin2019towards}, as one of the promising model compression techniques, reduces the complexity of CNNs by removing redundant parameters and computation from the three dimensions (depth, width, resolution) of CNNs. Specifically, width pruning \cite{molchanov2019importance, lin2020hrank, alwani2022decore, yu2022hessian, lin2019towards, kundu2021dnr, huang2022acceleration} devotes to compressing models by removing less important channels, while depth pruning \cite{wang2019dbp, lin2019towards, huang2018data, wen2016learning} conducts pruning at a coarser granularity (i.e., layer).
% which directly evaluate the contribution of each layer to the final results, and then remove less important layers. 
More recently, resolution pruning \cite{zhu2021dynamic, yang2020resolution, kong2022smart} has been proposed to compress the spatial redundancy in input images, which also effectively reduces the computational complexity of CNNs. However, the aforementioned approaches mainly focus on pruning a single dimension while ignoring the redundancy in the
other dimensions, which can only achieve a sub-optimal trade-off between model accuracy and execution efficiency. 

\begin{figure}
    \centering
    \includegraphics[width=0.48\textwidth]{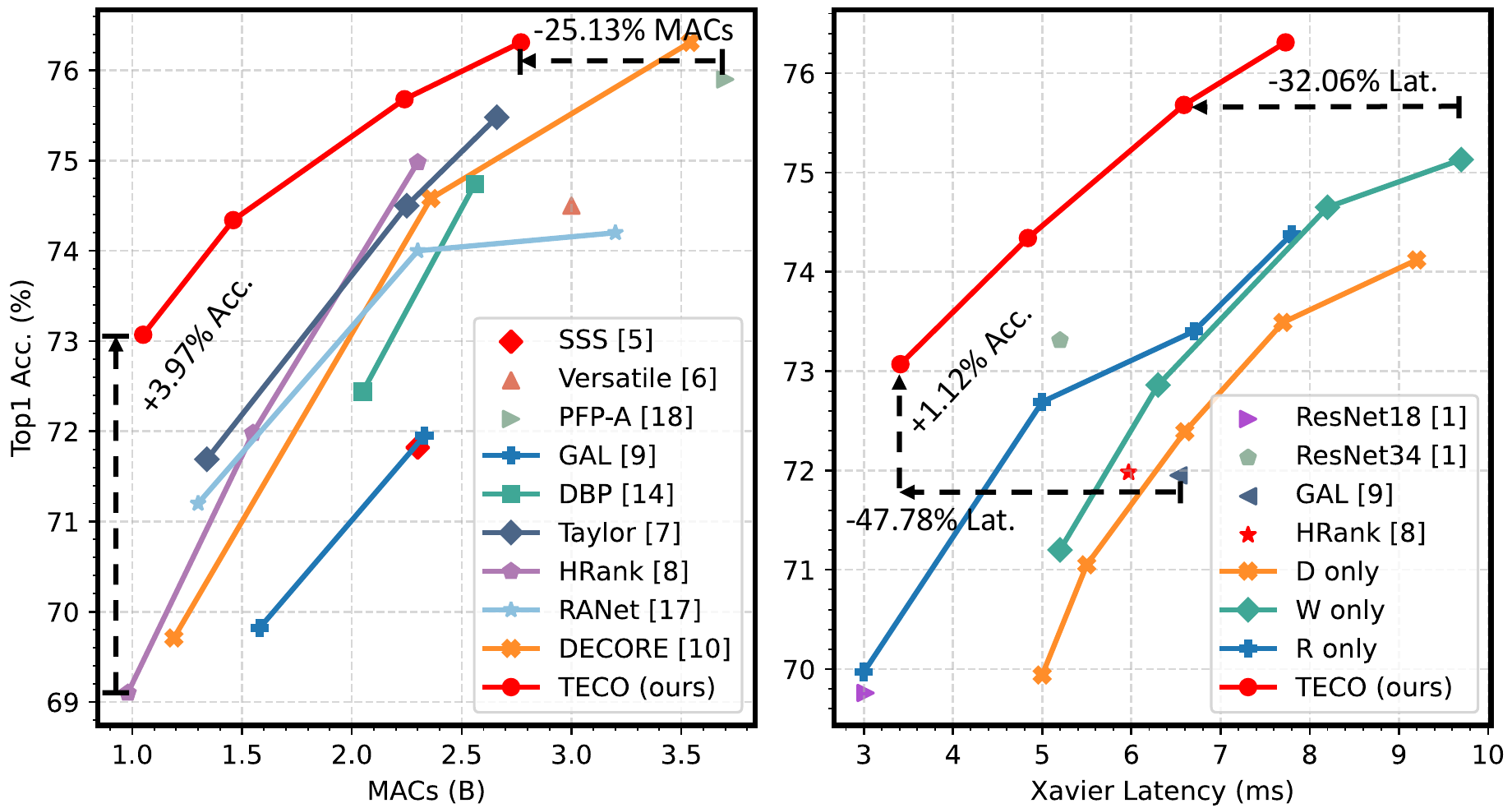}
    \caption{Comparison of different methods in terms of model MACs, accuracy, and inference latency. The baseline network is ResNet50 \cite{he2016deep}, which is trained on ImageNet. The latency is measured on AGX Xavier with a power budget of 30W.}
    \label{fig:introduction}
\end{figure}

\begin{figure*}
    \centering
    \includegraphics[width=0.98\textwidth]{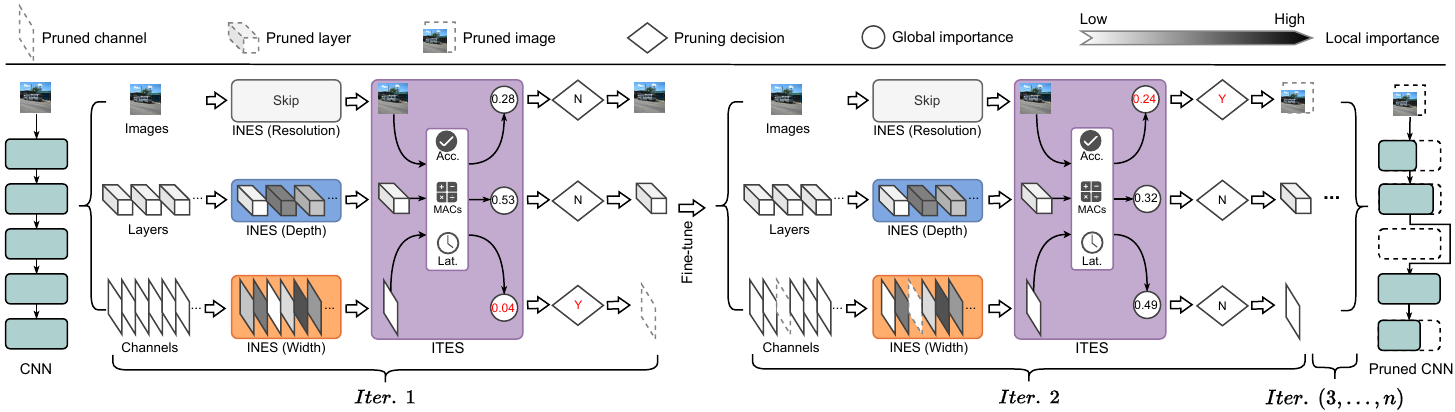}
    \caption{The overview of \mnade, where INES evaluates the local importance of units inside each dimension and ITES evaluates the global importance of units across different dimensions. The inner-dimensional evaluation is skipped for the resolution dimension (See Section \ref{sec:ines} for the detailed reason).}
    \label{fig:framework}
\end{figure*}

In this paper, we propose a multi-dimensional pruning framework, \mnade, to coordinately prune the three dimensions of CNNs. To accurately identify redundant units in the three dimensions, we first propose an inter-dimensional evaluation strategy (ITES) to comprehensively evaluate the importance of pruning units across different dimensions. In ITES, we integrate the contribution of each pruning unit to model complexity, accuracy, and inference latency into a unified metric, global importance, according to which the unit with the lowest global importance is considered redundant and can be safely removed.
% The first challenge faced by multi-dimensional pruning is to efficiently evaluate the importance of pruning units in the large design space formed by the three dimensions. Since the units of different dimensions (e.g., channel v.s. layer) are in different granularities, resulting in diverse impact on both model overhead and accuracy, they are unable to be fairly compared by existing evaluation strategies for single-dimensional pruning \cite{molchanov2019importance, lin2020hrank}. 
However, as ITES needs to collect multiple metrics for comprehensive evaluation, directly applying ITES to traverse all units of the three dimensions will be extremely time-consuming. To this end, we also introduce an inner-dimensional evaluation strategy (INES) to first quickly evaluate units within each dimension and identify the most redundant unit of each dimension. By this means, ITES only needs to be performed on the most redundant unit of each dimension for the final pruning decision. INES reduces the pruning candidates for each dimension from multiple to one, which reduces the evaluation cost of ITES and enhances the pruning efficiency significantly. 
On top of the two-step evaluation framework composed by INES and ITES, we design a heuristic pruning algorithm, which utilizes INES and ITES to progressively identify and prune redundant units. In this way, we can efficiently search for the optimal tiny architecture for resource-constrained embedded devices in the huge design space formed by the three dimensions.

% Based on ITES and INES, we design a heuristic pruning algorithm to progressively pruning the three dimensions, which efficiently optimizes the efficiency and accuracy of CNNs in the large design space formed by the three dimensions.

% Finally, we seamlessly integrate ITES and INES into \mnade, where the units will be first quickly evaluated inside each dimension, and the least important unit of each dimension will selected to perform ITES to determine the unit to prune. By this means, we are able to achieve both efficient and accurate multi-dimensional pruning.

% we further propose an inner-dimensional importance evaluation strategy to identify the redundant units inside each dimension, thereby achieve fine-grained pruning. Finally, we seamlessly integrate the inter-dimensional importance evaluation and inner-dimensional importance evaluation into a unified training process, which is capable of jointly pruning the three dimensions in an end-to-end and fully automated manner.

Our main contributions are three-fold:
\begin{enumerate}
    \item We introduce an inter-dimensional importance evaluation strategy (ITES) to evaluate the importance of units across different dimensions. We integrate the contribution of each unit to model complexity, accuracy and latency into a comprehensive metric, global importance, which enables us to accurately identify redundant units in the three dimensions.
    % \item We introduce a two-step importance evaluation framework, which contains a inter-dimensional
    \item We also propose an inner-dimensional importance evaluation strategy (INES) to quickly evaluate the local importance of units within each dimension, which reduces the pruning choices for each dimension from multiple to one, alleviating the evaluation overhead of ITES and improving the pruning efficiency of our framework.
    \item Based on ITES and INES, we design a heuristic pruning algorithm to progressively prune the three dimensions of CNNs. By iteratively identifying and removing redundant units with INES and ITES, our pruning algorithm can efficiently find the optimal tiny model for edge devices in the huge design space of multi-dimensional pruning.
    
    % \item \textcolor{red}{Based on ITES and INES, we propose a multi-dimensional pruning framework, \mnade, to efficiently identify and remove redundant units from the three dimensions, significantly optimizing the execution efficiency of CNNs on resource-constrained edge devices while maintaining high accuracy.}
\end{enumerate}

% We seamlessly integrate ITES and INES into \mnade, which executes ITES and INES iteratively to efficiently generate both accurate and fast model with any given resource budget.
As shown in Fig. \ref{fig:introduction}, our \mnade obtains 3.97\% higher top-1 accuracy on ImageNet than HRank \cite{lin2020hrank} with similar MACs.
% Meanwhile, compared to SSS \cite{wang2018learning}, we also achieve 1.25\% accuracy improvement with only 45\% MACs. 
For on-device acceleration, \mnade is 1.91$\times$ faster than GAL \cite{lin2019towards} while still achieving 1.12\% higher accuracy.

\section{Related Work}

\textbf{Single-dimensional pruning: } Computational redundancy widely exists in CNNs \cite{liu2021bringing}. To compress CNNs for a better trade-off between accuracy and execution efficiency, efforts have been made to remove redundancy from different dimensions of CNNs. Depth pruning \cite{he2016deep, wang2019dbp} devotes to compressing layer-level redundancy by pruning less important layers. Width pruning \cite{molchanov2019importance, lin2020hrank, Liebenwein2020Provable, alwani2022decore, yu2022hessian} conducts pruning at a finer granularity (i.e., channel), which yields compact models by removing channels with low sensitivity. Both depth pruning and width pruning focus on compressing the network architecture, while resolution pruning \cite{sandler2018mobilenetv2, zhu2021dynamic, yang2020resolution} optimizes the spatial redundancy in input images by shrinking images to a smaller resolution \cite{sandler2018mobilenetv2, zhu2021dynamic} or selectively cropping images for inference \cite{yang2020resolution}. However, all of the aforementioned pruning techniques only remove redundancy in a single dimension, which inevitably results in significant accuracy degradation as the compression rate increases.

\textbf{Multi-dimensional pruning: } Some works combine the pruning of different dimensions to achieve a higher compression rate while maintaining the accuracy. SSS \cite{huang2018data} and GAL \cite{lin2019towards} introduce additional regularization terms into the training process to learn a sparse architecture mask, according to which the layers and channels with higher sparsity will be removed. However, both approaches only combine the pruning of depth and width dimensions and fail to reduce the redundancy in input images, which loses the opportunity to further optimize the trade-off between model overhead and accuracy.

\textbf{Discussion: } Existing pruning frameworks mainly focus on compressing a single dimension or jointly compressing part of the three dimensions of CNNs. Instead, our method couples the compression of all dimensions, which is able to achieve a better result in terms of model efficiency and accuracy.

\section{Multi-Dimensional Pruning}

In this section, we first outline the design of \mnade, and then introduce each sub-component in detail.  

As demonstrated in Fig. \ref{fig:framework}, Given a CNN model $\mathcal{N}$, we first quickly evaluate the local importance of each unit inside each dimension with INES. Subsequently, the unit with the lowest local importance in each dimension is selected to perform ITES, where the three units of different granularities are fairly compared according to their global importance, and then the unit with the lowest global importance score is pruned safely. Afterwards, the model is fine-tuned to restore accuracy for the next pruning iteration. Finally, INES and ITES are executed iteratively to progressively prune the three dimensions of the given CNN model $\mathcal{N}$ for a better trade-off between model accuracy and execution efficiency.

% we propose an inter-dimension importance evaluation strategy (ITES), where the least important unit of each dimension is selected to perform a comprehensive cross-dimension comparison according to their impact on the accuracy and execution efficiency. Finally, by iteratively executing INES and ITES, we are capable of identifying and removing the redundancy from the three dimensions and achieving the optimal trade-off between accuracy and model efficiency.

\subsection{Inter-Dimensional Importance Evaluation (ITES)}

Compared to single-dimensional pruning \cite{lin2020hrank, wang2019dbp, yang2020resolution}, a major challenge faced by multi-dimensional pruning is to effectively evaluate and compare the importance of the pruning units of different dimensions. Intuitively, the pruning units of different dimensions (layer for depth, channel for width, pixel for resolution) are at different granularities, and pruning them can lead to diverse model complexity and accuracy. Moreover, for edge computing, pruning different dimensions also results in different run-time latency on embedded devices.
However, single-dimensional pruning \cite{molchanov2019importance, wang2019dbp, lin2020hrank} mainly considers model accuracy as the only metric to evaluate the importance of pruning units, which is inapplicable for evaluating units of different dimensions. To this end, we propose an inter-dimensional evaluation strategy (ITES) to comprehensively evaluate the importance of units across different dimensions according to their impacts on accuracy, model complexity, and on-device inference latency.

\begin{figure}[t]
    \centering
    \includegraphics[width=0.48\textwidth]{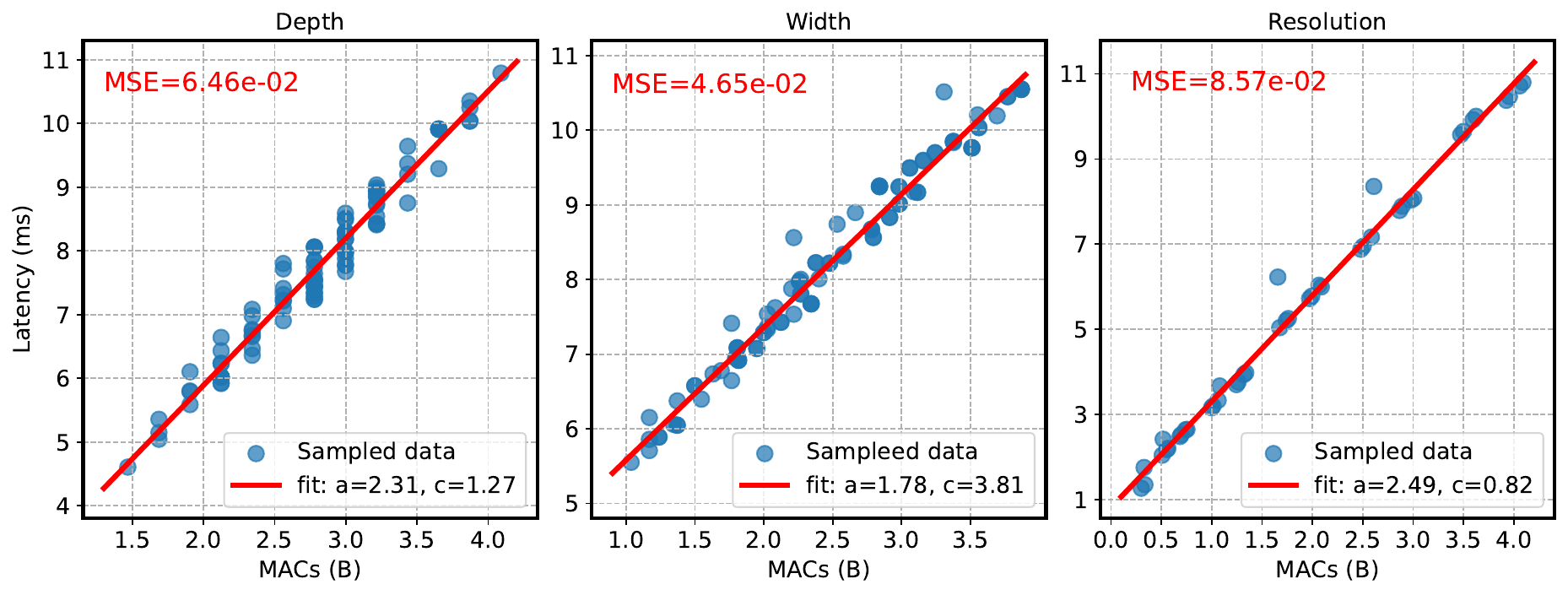}
    \caption{Latency distributions obtained by separately pruning the three dimensions. The low mean squared errors (MSE) reveal that the proposed latency model well fits the sampled data.}
    \label{fig:coeff}
\end{figure}

% As aforementioned, current model pruning approaches mainly focus on compressing a single dimension, which are only capable of evaluating and pruning the units of the same granularity. However, it is essential for multi-dimensional pruning to comprehensively evaluate the units of different granularity (e.g., channels v.s. layers) because pruning units of different granularity will result in different computation reduction and accuracy loss. Moreover, in our context (i.e., edge computing), pruning different dimensions can also have diverse impact on the inference latency on resource-constrained embedded devices. To this end, we propose an inter-dimensional importance evaluation strategy to comprehensively compare the overall importance of units of different granularity.
\textbf{Accuracy:} We quantify the contribution of a unit to accuracy as the increase in the cross entropy loss of the model prediction when removing this unit from the model. The cross entropy loss of image classification tasks can be defined as:
\begin{equation}
\label{eq:cel}
    L_{CE}\left(\mathcal{N}\right) = - \sum_{i=1}^n t_i \log(p_i)
\end{equation}
where $t_i$ is the ground truth probability for class $i$, $p_i$ is the predicted probability of model $\mathcal{N}$, and $n$ is the number of classes. Let $u$ be an arbitrary pruning unit of the three dimensions, 
% which can be a layer (for depth), a channel (for width) or a pixel (for resolution),
the impact of $u$ on accuracy is formulated as:
\begin{equation}
\label{eq:accuracy} 
\begin{split}
     \mathcal{A}(u) & = L_{CE}\left(\mathcal{N}'\right) - L_{CE}\left(\mathcal{N}\right) \\
     & = \sum_{i=1}^n t_i \log(p_i') - \sum_{i=1}^n t_i \log(p_i) = \sum_{i=1}^n t_i \log(\frac{p_i'}{p_i})
\end{split}
\end{equation}
where $\mathcal{N}'$ is the pruned model by removing $u$ from $\mathcal{N}$.

\textbf{Model Complexity:} We use model MACs to quantify the complexity of CNNs as all three dimensions can affect model MACs while parameters are only related to depth and width. Therefore, the impact of $u$ on model complexity can be efficiently measured by calculating the MACs reduction achieved by removing $u$, which is represented as:
\begin{equation}
    \label{eq:macs}
    \mathcal{M}(u) = \left| \text{MACs}(\mathcal{N}') - \text{MACs}(\mathcal{N}) \right|
\end{equation}

\textbf{On-Device Latency:} The most intuitive way to evaluate the impact of $u$ on latency is to measure the latency reduction. However, directly measuring the latency reduction on embedded devices during pruning will incur a huge time cost and reduce the pruning efficiency. To this end, we introduce a dimension-wise latency model to efficiently estimate the latency reduction according to the MACs reduction of each dimension. To build the dimension-wise latency model, we vary the three dimensions of ResNet50 to sample models with different MACs and measure their latency on the target device. The sampling results are summarized in Fig. \ref{fig:coeff}, which demonstrates a linear relationship between the latency and MACs. Therefore, we formulate the latency model for each dimension as follows:
% we introduce a dimension-wise acceleration factor to quantify the impact of pruning different dimensions on latency. The acceleration factor of a dimension is defined as the speed at which the latency decreases when pruning that dimension. To calculate the acceleration factor, we separately sample multiple models with different MACs along the three dimensions and measure their latency on the target device. Thereafter, we model the relationship between model MACs and latency with a simple yet effective linear model:
\begin{equation}
    \label{eq:latency}
    l_s = a_s \cdot m + c_s
\end{equation}
where $l_s$ is the predicted latency for dimension $s \in \{d, w, r\}$ and $m$ is the model MACs. $a_s$ and $c_s$ are dimension-wise hyperparameters, which are fitted by Least Squares Method with the sampled data. Since the residual network architecture of ResNet50 is widely utilized in many advanced CNNs, the established latency model can be well generalized to other advanced CNNs. Let $s$ be the dimension of unit $u$, the impact of $u$ on latency (i.e., the latency reduction) is derived as:
\begin{equation}
    \label{eq:latency_impact}
    \mathcal{T}(u) = |l_s' - l_s| = a_s(\left| m' - m \right|) = a_s \mathcal{M}(u)
\end{equation}
where $m'$ and $l_s'$ are the MACs and estimated latency of the pruned model $\mathcal{N}'$. Through Equation \ref{eq:latency_impact}, we can quickly evaluate the impact of $u$ on latency using the MACs reduction and the dimension-wise latency hyperparameter $a_s$. As shown in Fig. \ref{fig:coeff}, the value of the latency hyperparameter $a_s$ varies across dimensions, which reveals that, for different dimensions, the same reduction on MACs can lead to diverse latency reduction.

% After determining the latency model of each dimension, the acceleration factor can be obtained by calculating the gradient of $l_s$, which is represented as:
% \begin{equation}
    % \label{eq:gradient}
    % g_s = \frac{\mathrm{d} l_s}{\mathrm{d} m} = a_s
% \end{equation}

\textbf{Global Importance:} Finally, we combine the impact of $u$ on accuracy, model complexity, and latency as a unified metric coined global importance, which is formulated as:
\begin{equation}
    \label{eq:inter}
    \mathcal{I}(u) = \frac{\mathcal{A}(u)}{\alpha \mathcal{M}(u) + (1-\alpha)\mathcal{T}(u)}
\end{equation}
where $\alpha \in [0, 1]$ is a hyperparameter to control the contribution of $\mathcal{M}(u)$ and $\mathcal{T}(u)$, which provides ITES with enough flexibility to accommodate different design considerations. Specifically, by increasing the value of $\alpha$, ITES will focus more on the reduction of model complexity (i.e., MACs). Otherwise, the on-device latency will be the main consideration of ITES. In our experiments, we empirically set $\alpha=0.5$ to equalize their contribution to the global importance. 
According to Equation \ref{eq:inter}, unit $u$ is considered less important if pruning it can bring more significant reduction in model MACs and latency with less increase in prediction loss.

% To yield compressed models that can be efficiently executed on resource-constrained edge devices, we also incorporate the impact of different dimensions on latency into our evaluation strategy. First, we separately sample multiple models with different MACs along the three dimensions through random single-dimensional pruning, and then we deploy these sampled models onto the target edge device to evaluate their actual inference latency. The relationships between MACs and latency of different dimensions are summaried in Fig. \ref{fig:coeff}, where we can see that the latency decreases at different speeds for different dimensions. First, we model the latency reduction along MACs with a simple yet effective linear model, which can be formulated as follows:
% \begin{equation}
%     \label{eq:latency}
%     l_s = g_s \cdot m + c_s
% \end{equation}
% where $l_s$ denotes the predicted inference latency for dimension $s\in \{d, w, r\}$, $m$ is the MACs of the pruned model, $g_s$ and $c_s$ are the hyper-parameters to fit. As demonstrated in Fig. \ref{fig:coeff}, the proposed linear model can well fit sampled data. 

\subsection{Inner-Dimensional Importance Evaluation (INES)}
\label{sec:ines}

\begin{figure}
    \centering
    \includegraphics[width=0.48\textwidth]{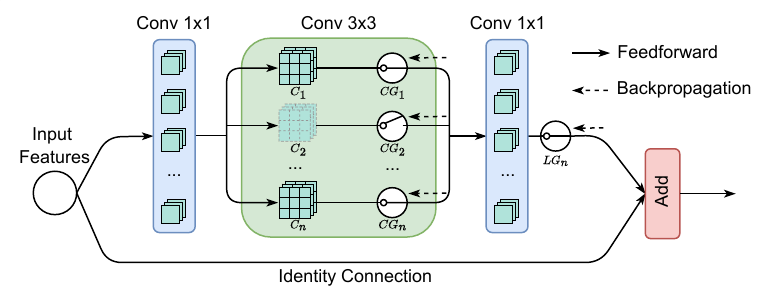}
    \caption{The architecture of the fully gated residual bottleneck block with channel gates and a layer gate.}
    \label{fig:block}
\end{figure}

ITES effectively evaluates units across different dimensions using global importance. However, obtaining the global importance of a unit is relatively time-consuming due to the calculation of $\mathcal{A}(u)$, $\mathcal{M}(u)$, and $\mathcal{T}(u)$, and thus directly using ITES to traverse the three dimensions will lead to an expensive evaluation cost, thereby degrading the efficiency of our pruning framework. Therefore, we further propose an inner-dimensional evaluation strategy (INES) to cooperate with ITES to efficiently evaluate all units. 
INES first quickly evaluates the local importance of units inside each dimension, and then only the unit with the lowest local importance in each dimension will be selected to perform ITES for its global importance. In this way, ITES will be performed only on three units and thus the evaluation overhead is reduced significantly. Finally, the unit with the lowest global importance will be considered redundant and removed. Through such a two-step evaluation mechanism, we are capable of accurately and efficiently identifying the redundant units in the three dimensions.

% We further propose an inner-dimensional evaluation strategy (INES) to first quickly evaluate units inside each dimension with a metric namely local importance. 
% According to the local importance, only the unit with the lowest local importance of each dimension will be selected to perform ITES for its global importance. Finally, the unit with the lowest global importance will be pruned.
% By such a two-step evaluation mechanism, the algorithm complexity of evaluation is greatly reduced to $\mathcal{O}(1)$.

To efficiently evaluate the local importance of units inside each dimension, we design a fully gated residual bottleneck block. The block architecture is demonstrated in Fig. \ref{fig:block}, where each channel is followed by a channel gate ($CG$). Meanwhile, there is a layer gate ($LG$) at the end of the block. These gates are introduced for two reasons: (1) to control the pruning of each channel or the whole layer by setting the gate's weight to 0 (pruned) or 1 (preserved); (2) to quickly calculate the local importance for inner-dimensional evaluation.

\textbf{Local Importance:} Inspired by the channel pruning approach proposed in \cite{molchanov2019importance}, we approximate the local importance of a channel with the gradient of the corresponding channel gate, which can be formulated as follows:
\begin{equation}
    \label{eq:channel_importance}
    \mathcal{I}^w_i = \left( \frac{\partial L_{CE}}{\partial CG_i} \right)^2
\end{equation}
where $\mathcal{I}^w_i$ is the local importance of the $i$-th channel and $CG_i$ is the gate of the $i$-th channel. 
% By this means, the inner-dimension importance of all channels can be efficiently calculated through back-propagation.
Further, we extend this idea to the depth dimension, where we add a layer gate at the end of each layer to collect the layer-level gradients and utilize the layer-level gradients to quantify the importance of each layer:
\begin{equation}
    \label{eq:layer_importance}
    \mathcal{I}^d_i = \left( \frac{\partial L_{CE}}{\partial LG_i} \right)^2
\end{equation}
where $\mathcal{I}^d_i$ and $LG_i$ represent the local importance and the gate of the $i$-th layer, respectively. Thanks to our fully gated block architecture, we are able to simultaneously obtain the gradient of all channel gates and layer gates by only performing backpropagation once, and thus the evaluation cost of INES is reduced significantly. 
In practise, we randomly select multiple images from the training set to perform backpropagation and average their gradients to obtain a more consistent and accurate estimation for the local importance of each unit. In our test, we empirically observe that 5,000 images are adequate to produce an accurate estimation. Using more images brings only negligible accuracy improvement at the expense of larger time overhead, degrading the efficiency of our approach. The backpropagation of 5,000 images only takes about 2.94 seconds on a RTX3090 GPU, which validates the efficiency of INES.

For the resolution dimension, selectively pruning the millions of images in large-scale datasets (e.g., ImageNet) is unpractical for our framework due to the enormous overhead \cite{zhu2021dynamic}. Instead, we implement resolution pruning by uniformly shrinking all images, which eliminates the evaluation cost of INES for the resolution dimension, enhancing the efficiency of our method.

% To efficiently calculate the inter-dimension importance, we introduce a fully-gated block architecture, which is demonstrated in Fig. \ref{fig:block}. The $CG_i$ is the gate for the $i$-th channel, while $LG_i$ is the gate for the $i$-th layer. Inspired by \cite{molchanov2019importance}, the importance of a parameter can be quantified as the change of the prediction error when removing the parameter ($z_m$):
% \begin{equation}
%     \label{eq:importance}
%     \mathcal{I}_m = \left( E(\mathcal{D}, Z) - E(\mathcal{D}, Z|z_m=0) \right)^2
% \end{equation}
% where $\mathcal{D}$ denotes the evaluation dataset and $Z$ is the collection of all parameters. However, to measure the importance of millions of parameters with Equation \ref{eq:importance} will be computationally prohibitive. To address this problem, $\mathcal{I}_m$ can be approximated by the first-order Taylor expansion:
% \begin{equation}
% \label{eq:fo-expansion}
%     \mathcal{I}_m(Z) = \left(g_mz_m\right)^2
% \end{equation}
% where $g_m$ denotes the gradient of the parameter $z_m$, which can be efficiently obtained through back-propagation. After obtaining the importance of all parameters, the importance of a channel can be seen as the group contribution of the parameters in this channel, which is approximated by the gradient of the corresponding channel gate:
% \begin{equation}
%     \label{eq:gate-gradient}
%     \mathcal{I}_{S_i}(Z) = \left( \frac{\partial E}{\partial CG_i} \right)^2 = \left(\sum_{s\in S_i} g_sz_s \right) ^2
% \end{equation}
% where $S_i$ represents the $i$-th channel and $s$ is the parameter in $S_i$. 

\subsection{Heuristic Architecture Descent}
\label{subsec:iterative}

\begin{algorithm}[t]
\small
\caption{Heuristic Architecture Descent}\label{alg:one}
\KwData{overparameterized CNN $\mathcal{N}$, training dataset $D_t$, evaluation dataset $D_v$, pruning iterations $n$ 
% channel pruning stride $s_w$, layer pruning stride $s_d$, resolution pruning stride $s_r$
}
\KwResult{pruned network $\mathcal{N}_p$}
$iter \gets 0$\;
\While{$iter < n$}{
    \Comment{\footnotesize Inner-dimensional evaluation}
    $(d^*, w^*, r^*) \gets \text{INES}(\mathcal{N}, D_v)$ \\
    % $d^* \gets \text{INES}_{layer}(\mathcal{N}, D_v) $  \Comment{\footnotesize Evaluate layers} 
    % $r^* \gets \text{INES}_{resolution}(\mathcal{N}, D_v)$ \Comment{\footnotesize For resolution} 
    \Comment{\footnotesize Inter-dimensional evaluation}
    $dim \gets \text{ITES}(\mathcal{N}, D_v, d^*, w^*, r^*)$ \\
    \uIf{$dim$ is depth}{
        Prune($\mathcal{N}, d^*$) \Comment{\footnotesize Prune the layer}
    } 
    \uElseIf{$dim$ is width}{
        Prune($\mathcal{N}, w^*$) \Comment{\footnotesize Prune the channel}
    }
    \ElseIf{$dim$ is resolution}{
        Prune($\mathcal{N}, r^*$) \Comment{\footnotesize Prune the image}
    }
    Fine-tune($\mathcal{N}$, $D_t$) \\
    $iter \gets iter + 1$
}
$\mathcal{N}_p \gets \mathcal{N}$
\end{algorithm}

To efficiently find the optimal architecture for a given resource budget, we further propose a heuristic pruning algorithm, which progressively executes INES and ITES to prune redundant units from the three dimensions. Inspired by gradient descent, we coin this algorithm heuristic architecture descent as the architecture is gradually descending along the direction that achieves the best efficiency-accuracy trade-off. 

The proposed heuristic architecture descent is defined in Algorithm \ref{alg:one}. Given a baseline network $\mathcal{N}$, a training dataset $D_t$, and an evaluation dataset $D_v$ that consists of 5,000 randomly selected images from $D_t$, we perform the pruning operation for $n$ iterations. For each iteration, we first conduct INES for $\mathcal{N}$ on $D_v$ to obtain the local importance of each unit within each dimension. Based on the local importance, we select the least important unit of each dimension ($d^*$ for depth, $w^*$ for width, and $r^*$ for resolution) to perform ITES for their global importance, according to which the unit with the lowest global importance score will be pruned. Subsequently, the model is fine-tuned on $D_t$ for 1 epoch to retain its accuracy for the next pruning iteration, where the optimizer for fine-tuning is SGD with a learning rate of 1e-4. The completely pruned model $\mathcal{N}_p$ will be generated after $n$ iterations. The value of $n$ depends on the resource budget. The smaller the budget, the larger the value of $n$. In practise, 10 iterations are adequate to obtain a compact model, which indicates that the time cost of pruning is much smaller than the main training of CNNs. Finally, the completely pruned model will be trained from scratch as described in Section \ref{subsec:imgnet} for the final accuracy. Compared to global search algorithms \cite{huai2021zerobn} that directly search the huge design space for the optimal solution, our heuristic pruning algorithm greatly reduces the exploration overhead.

% To leverage the pruning cost and the performance of the pruned model, each dimension will be pruned with a given pruning stride, which indicates how many units will be remove for one iteration. In practice, we set the channel pruning stride $s_w=50$, the layer pruning stride $s_d=1$, and the resolution pruning stride $s_r=16$. 

% At the end of each iteration, the pruned model will be fine-tuned for 1 epoch on the training dataset $D_t$ to recover the accuracy for the next iteration. 

% \textbf{Fine-tuning v.s. Training from scratch:} There are two schemes to train the pruned model for the final accuracy: (1) training from scratch; (2) inheriting and fine-tuning the weights from the baseline model. As discussed in \cite{liu2019rethinking}, fine-tuning only achieves comparable or even worse accuracy than training from scratch. Therefore, we train the pruned model from scratch.

% \begin{equation}
%     g_s = \frac{\mathrm{d} l_s}{\mathrm{d} x}
% \end{equation}

% \begin{equation}
%     c_s = \frac{g_s}{\min (g_d, g_w, g_r)}
% \end{equation}
% where $s \in \{d, w, r\}$ denotes the dimension. $c_d=1.61, c_w=1.00, c_r=1.93$
\section{Experiments}
\label{sec:experiments}

% \begin{table}[bp]
%   \centering
%   \caption{The specifications of utilized hardware platforms.}
%     \resizebox{0.48\textwidth}{!}{
%     \begin{tabular}{l||r||c||r||c||r}
%     \toprule
%     \toprule
%     \textbf{Device} & \textbf{Power} & \textbf{CPU Freq.} & \textbf{GPU Freq.} & \textbf{CUDA Cores} &  \textbf{Memory} \\
%     \midrule
%     % Raspberry Pi 4 & 6 W & 1536 MHz & N.A. & N.A. & 4 GB \\
%     Jetson Nano     & 10 W    & 1479 MHz & 922 MHz & 128 & 4 GB \\
%     Jetson TX2    & 15 W & 1400 MHz & 1122 MHz & 256 &  8 GB\\
%     AGX Xavier     & 30 W    & 1780 MHz & 900 MHz & 512 & 16 GB \\
%     \bottomrule
%     \bottomrule
%     \end{tabular}%
%     }
%   \label{tab:spec}%
% \end{table}%

\begin{table}[tbp]
  \centering
  \caption{Comparison with SOTA pruning approaches on ImageNet. The baseline network is ResNet50. \{d, w, r\} indicate the pruned dimensions in different methods.}
    \resizebox{0.48\textwidth}{!}{
    \begin{tabular}{lcccrrr}
    \toprule
    \toprule
    \textbf{Method} & \textbf{d} & \textbf{w} & \textbf{r} & \textbf{MACs (B)} & \textbf{Acc@1 (\%)} & \textbf{Acc@5 (\%)} \\
    \midrule
    ResNet50 \cite{he2016deep} &       &       &       & 4.10 & 76.80 & 93.38  \\
    \midrule
    DECORE-4 \cite{alwani2022decore} &       & \checkmark     &       & 1.19  & 69.71 & 89.37 \\
    ResNet18 \cite{he2016deep} & \checkmark     &       &       & 1.80  & 69.76 & 89.08 \\
    GAL-1 \cite{lin2019towards}& \checkmark     & \checkmark     &       & 1.58 & 69.82 & 89.75 \\
    Bilinear \cite{sandler2018mobilenetv2} &       &       & \checkmark     & 1.10 & 69.97 & 89.19 \\
    HAP \cite{yu2022hessian} &       & \checkmark     &       & 1.34  & 71.18 & - \\
    Taylor \cite{molchanov2019importance} &       & \checkmark     &       & 1.34  & 71.69 & - \\
    HRank \cite{lin2020hrank} &       & \checkmark     &       & 1.55 & 71.98 & 91.09 \\
    DBP-0.5 \cite{wang2019dbp} & \checkmark     &       &       & 2.05  & 72.44 & - \\
    \textbf{\mnade-S (ours)} & \boldcheckmark & \boldcheckmark & \boldcheckmark &    \textbf{1.05}   &  \textbf{73.07} & \textbf{91.18}  \\
    \midrule
    SSS-26 \cite{huang2018data} & \checkmark     & \checkmark     &       & 2.33  &  71.82 & 90.79  \\
    GAL-0.5 \cite{lin2019towards} & \checkmark     & \checkmark     &       & 2.33   & 71.95 & 90.94 \\
    ResNet34 \cite{he2016deep} & \checkmark     &       &       & 3.70   & 73.31 & 91.42 \\
    Bilinear \cite{sandler2018mobilenetv2} &       &       & \checkmark     & 2.53   & 73.40 & 91.30 \\
    % HAP \cite{yu2022hessian}   &       & \checkmark     &       & \textbf{1.67} & \textbf{8.88} & 74.00 \\
    RANet \cite{yang2020resolution} &       &       & \checkmark     & 2.30  & 74.00 & - \\
    Taylor \cite{molchanov2019importance} &       & \checkmark     &       & 2.25   & 74.50 & - \\
    % DECORE-6 \cite{alwani2022decore} &       & \checkmark     &       & 2.36   & 74.58 & 92.18 \\
    DBP-0.4 \cite{wang2019dbp} & \checkmark     &       &       & 2.56     & 74.74 & - \\
    HRank \cite{lin2020hrank} &       & \checkmark     &       & 2.30   & 74.98 &  92.33 \\
    % Adapt-DCP \cite{liu2021discrimination} &       & \checkmark     &       & \textbf{1.95}   & 75.15 & 92.30 \\
    DR-ResNet50 \cite{zhu2021dynamic} &       &       & \checkmark     & 2.30   & 75.30 & 92.20 \\
    \textbf{\mnade-M (ours)} & \boldcheckmark & \boldcheckmark & \boldcheckmark &  \textbf{2.24}  & \textbf{75.68} & \textbf{92.79} \\
    \midrule
    SSS-32 \cite{huang2018data} & \checkmark     & \checkmark     &       & 2.82  & 74.18 & 91.91 \\
    Bilinear \cite{sandler2018mobilenetv2} &       &       & \checkmark     & 3.00  & 74.30 & 91.90 \\
    % AutoPruner \cite{luo2020autopruner} &       & \checkmark     &       & 3.76    & 74.76 & 92.15 \\
    HAP \cite{yu2022hessian}  &       & \checkmark     &       & 2.71   & 75.12 & - \\
    C-SGD70 \cite{ding2019centripetal} &       & \checkmark     &       & \textbf{2.60}     & 75.30 & 92.50 \\
    Taylor \cite{molchanov2019importance} &       & \checkmark     &       & 2.66  & 75.48 & - \\
    PFP-A \cite{Liebenwein2020Provable} &       & \checkmark     &       & 3.70   & 75.90 & 92.80 \\
    % DR-ResNet50 \cite{zhu2021dynamic} &       &       & \checkmark     & 2.70  & 30.50 & 76.20 \\
    DECORE-8 \cite{alwani2022decore} &       & \checkmark     &       & 3.54   & 76.31 & 93.02 \\
    \textbf{\mnade-L (ours)} & \boldcheckmark & \boldcheckmark & \boldcheckmark &  2.87      & \textbf{76.34} & \textbf{93.20}  \\
    \bottomrule
    \bottomrule
    \end{tabular}%
    }
  \label{tab:imgnet}%
\end{table}%

In this section, we conduct extensive experiments to validate the superiority of \mnade over other SOTA approaches in terms of accuracy, model complexity, and run-time latency. In our experiments, we select three widely utilized embedded platforms: 1) Jetson Nano, 2) Jetson TX2, and 3) AGX Xavier, to deploy pruned models obtained from different approaches and measure their actual inference latency. Also, we perform ablation experiments to validate the efficacy of each component.

\subsection{Experiments on ImageNet}
\label{subsec:imgnet}
\textbf{Setup: } On the most representative large-scale dataset, ImageNet, we train all models from scratch for 120 epochs using SGD optimizer with a momentum of 0.9. The batch size for training is 1024. Correspondingly, the initial learning rate is set to 1.6 and decayed by cosine annealing scheduling \cite{loshchilov2017sgdr}. In addition, the learning rate for fine-tuning is 1e-3. To prevent overfitting, we also use label smoothing with $\epsilon=0.1$.

The results are summarized in TABLE \ref{tab:imgnet}, which shows that our approach achieves the highest accuracy across a wide spectrum of model MACs. Specifically, in the low compute regime, our \mnade-S achieves $3.36\%$ higher top-1 accuracy with about $12\%$ less MACs compared to DECORE-4 \cite{alwani2022decore}. In comparison with GAL-1\cite{lin2019towards}, \mnade-S also improves the top-1 accuracy by $3.25\%$ while reducing the MACs by $33.5\%$. In the highest MACs regime, \mnade-L outperforms SSS-32 \cite{huang2018data} with $2.16\%$ higher top-1 accuracy. 
% This experiment reveals that our approach significantly reduces the model complexity without sacrificing accuracy.

% TABLE generated by Excel2LaTeX from sheet 'Sheet1'

\begin{figure}[tbp]
    \centering
    \includegraphics[width=0.48\textwidth]{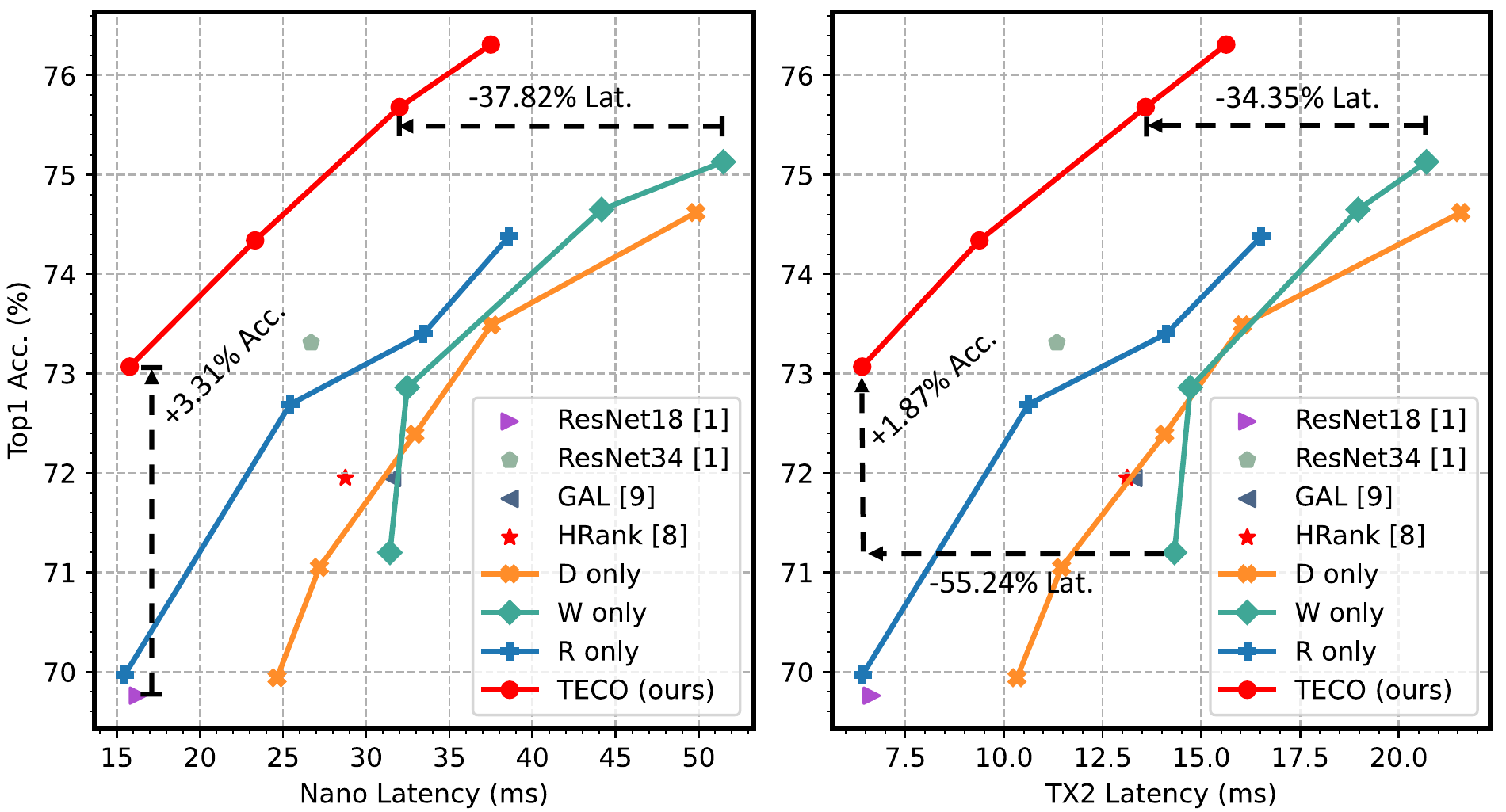}
    \caption{Comparison of inference latency on Jetson Nano and Jetson TX2. For the consistency of results, all models are executed for 30 iterations to get the average inference latency.}
    \label{fig:latency}
\end{figure}

\subsection{Comparison of On-Device Acceleration}
\label{subsec:on-device}

In this experiment, we evaluate the run-time latency of models pruned by different frameworks on three widely used edge platforms: AGX Xaiver, Jetson TX2, and Jetson Nano. The corresponding results are shown in Fig. \ref{fig:introduction} and Fig. \ref{fig:latency}, respectively. We observe that our approach surpasses all competitors on all devices. Specifically, our method achieves $3.7\%$ higher accuracy (75.68\% v.s. 71.98\%) than HRank \cite{lin2020hrank} with a similar latency (13.59 ms v.s. 13.12 ms) on Jetson TX2. On Jetson Nano, a more resource-economic edge device, our \mnade-S observes 1.12\% higher accuracy (73.07\% v.s. 71.95\%) than GAL \cite{lin2019towards} with only 50\% latency budget (15.78 ms v.s. 31.56 ms). Similar results are also observed on Xavier. 
% This experiment proves that that our framework can greatly optimize the on-device efficiency of CNNs on various embedded devices.
The experiment validates the efficacy of our method in optimizing the execution efficiency of CNNs on various embedded devices.
% This experiment on latency validates that our method remarkably optimizes the execution efficiency on various edge devices while maintaining high accuracy.

\subsection{Experiments on CIFAR-10}
\label{subsec:cifar}

To validate the efficacy of \mnade in extremely resource-constrained environments (e.g., TinyML), we conduct experiments to further compress small models for edge devices.

\begin{table}[htbp]
  \centering
  \caption{Experiments on CIFAR-10. "-" means no source code is provided for reproducing the experiment.}
    \resizebox{0.48\textwidth}{!}{
    \begin{tabular}{lrrrr}
    \toprule
    \toprule
    \textbf{Method} & \multicolumn{1}{c}{\textbf{MACs (M)}} & \multicolumn{1}{c}{\textbf{Params (M)}} & \textbf{Latency (ms)} & \textbf{Acc@1 (\%)} \\
    \midrule
    ResNet110 \cite{he2016deep} & 252.89 & 1.72  & 4.98  & 93.50 \\
    \midrule
    GAL-0.5 \cite{lin2019towards} & 130.20 & 0.95  & 2.96  & 92.55 \\
    HRank-2 \cite{lin2020hrank} & \textbf{79.30} & \textbf{0.53}  & 3.49  & 92.65 \\
    HRank-1 \cite{lin2020hrank} & 105.70 & 0.70  & 3.92  & 93.36 \\
    DECORE-300 \cite{alwani2022decore} & 96.66 & 0.61  & -     & 93.50 \\
    DECORE-500 \cite{alwani2022decore} & 163.30 & 1.11  & -     & 93.88 \\
    \textbf{\mnade-Tiny (ours)} &   108.60    &   0.98    &  \textbf{2.94}     &  \textbf{93.94} \\
    \bottomrule
    \bottomrule
    \end{tabular}%
    }
  \label{tab:cifar10}%
\end{table}%

\textbf{Setup: } We use ResNet110 \cite{he2016deep} as the baseline network and use CIFAR-10 as the dataset. ResNet110 is a lightweight CNN specially designed for tiny images, which only contains 1.7M parameters and 252M MACs. All models are trained for 200 epochs using SGD optimizer. The batch size is 128 and the initial learning rate is 0.1, which is decayed by cosine annealing \cite{loshchilov2017sgdr}. The latency of all models are measured on Jetson Nano.

% TABLE generated by Excel2LaTeX from sheet 'Sheet1'

The results in TABLE \ref{tab:cifar10} show that our method achieves the best latency-accuracy trade-off. For instance, compared to HRank-2 \cite{lin2020hrank}, we achieve 1.29\% higher accuracy with only 84.2\% inference latency. Interestingly, our method reduces the MACs of the baseline ResNet110 by 57\% while still achieving 0.44\% higher accuracy, which is because CNNs usually overfit on CIFAR-10 \cite{alwani2022decore}, and our approach greatly mitigates the overfitting by comprehensively reducing the redundancy in the three dimensions, thereby improving the accuracy.

\subsection{Ablation Study \vspace{5pt}}
\label{subsec:ablation}
% We introduce two novel components, INES and ITES in \mnade to enable efficient and accurate multi-dimensional pruning. To validate the efficacy of each component, we perform comprehensive ablation experiments on ImageNet. First, we perform random multi-dimensional pruning by removing both INES and ITES from \mnade. The results in TABLE \ref{tab:ablation} show that, without INES and ITES, the multi-dimensional pruning only achieves comparable accuracy to single-dimensional pruning. Subsequently, we separately retrieve INES and ITES, and both of which observe a remarkable accuracy gain. Finally, the complete \mnade framework with both INES and ITES achieves the best accuracy with the lowest latency. Specifically, the complete \mnade observes 1.01\% accuracy improvement and 11.3\% latency reduction compared to the random multi-dimensional pruning. The ablation experiment indicates that both INES and ITES contribute the final performance.
We introduce two novel evaluation strategies, INES and ITES in \mnade to enable efficient and accurate multi-dimensional pruning. To validate the efficacy and efficiency of each component, we perform comprehensive ablation experiments on ImageNet. First, we perform multi-dimensional pruning with both INES and ITES removed from \mnade. The results in TABLE \ref{tab:ablation} show that, without INES and ITES, the multi-dimensional pruning only achieves comparable accuracy to single-dimensional pruning. It is worth noting that, for resolution pruning, we directly shrink the resolution of images and do not evaluate the pruning units of the network, thus the pruning cost is negligible. Then, we separately retrieve INES and ITES, and both of which observe a remarkable accuracy gain. When only using ITES to evaluate all units, the pruned model achieves the best performance in terms of MACs, inference latency, and accuracy, which proves that ITES is able to accurately identify the redundancy in the three dimensions. However, this strategy also results in an unbearable time cost. In contrast, by collaboratively using INES and ITES, we achieve competitive model performance while reducing the pruning cost by 83.21\% compared to using ITES alone, which greatly increases the efficiency of our framework. The ablation study reveals that both INES and ITES contribute to our pruning framework.

\begin{figure}[!t]
    \centering
    \includegraphics[width=0.48\textwidth]{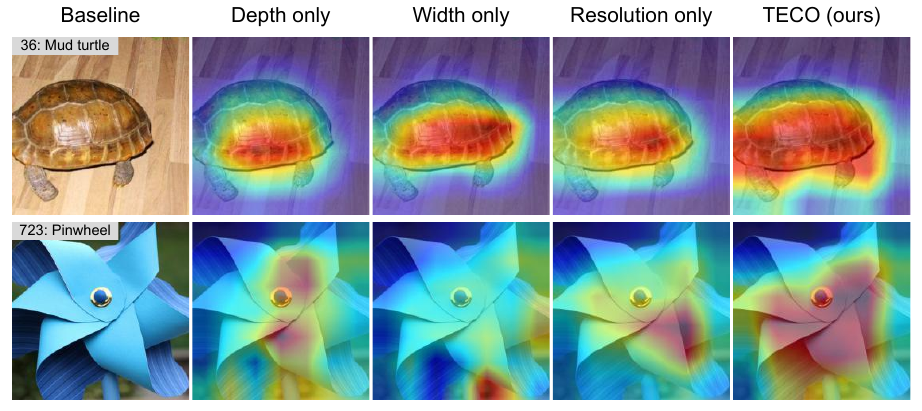}
    \caption{The class activation map (CAM) for different pruning methods. The region in red is the most contributing part of the image. The images are randomly selected from ImageNet.}
    \label{fig:cam}
\end{figure}

% TABLE generated by Excel2LaTeX from sheet 'Sheet1'
\begin{table}[htbp]
  \centering
  \caption{The results of ablation experiments. The baseline network is ResNet50 and the latency is measured on Xavier. We quantify the pruning cost of different methods as the time consumed on a single RTX3090 GPU for pruning.}
    \resizebox{0.48\textwidth}{!}{
    \begin{tabular}{lrrrr}
    \toprule
    \toprule
    \textbf{Method} & \textbf{MACs (B)} & \textbf{Latency (ms)} & \textbf{Acc@1 (\%)} & \textbf{Cost (hour)} \\
    \midrule
    ResNet50 & 4.10  & 10.60 & 76.80 & - \\
    \midrule
    Depth only &  2.30     &     6.60  &    72.39   & 7.52  \\
    Width only &    2.30   &    8.20   &     74.65  & 26.44  \\
    Resolution only &     2.50  &     6.70  &    73.40   & 0.00  \\
    w/o INES + w/o ITES &   2.17    &  7.34     & 74.67    & 20.77  \\
    INES + w/o ITES &   2.30    &  6.84     &  75.28     & 21.26  \\
    w/o INES + ITES &   \textbf{2.12}    &  \textbf{6.42}     &   \textbf{75.71}    & 179.42  \\
    \textbf{Ours (INES + ITES)} &   2.24    &  6.51     &   75.68    &  30.12\\
    \bottomrule
    \bottomrule
    \end{tabular}%
    }
  \label{tab:ablation}%
\end{table}%

\subsection{Interpretability Analysis\vspace{5pt}}
\label{subsec:cam}

To gain insight into the advantages of our approach, we visualize the class activation map \cite{selvaraju2017grad} for \mnade and single-dimensional pruning. The results are demonstrated in Fig. \ref{fig:cam}, where we observe that single-dimensional pruning approaches only focus on part of the foreground object of input images, which may overlook important features and consequently lead to wrong predictions. In contrast, our \mnade utilizes the whole foreground object for prediction, which effectively addresses the aforementioned problem of single-dimensional pruning, significantly improving model accuracy.

\vspace{5pt}
\section{Conclusion\vspace{5pt}}

\label{sec:conclusion}

% In this paper, we propose a multi-dimensional pruning framework, \mnade, to jointly prune the three dimensions of CNNs for resource-constrained embedded devices.
% First, we introduce an inter-dimensional evaluation strategy (ITES), which quantifies the impacts of a unit on model computation, accuracy and on-device latency into a novel metric, global importance, by which we achieve a comprehensive and fair comparison of units across different dimensions. 
% Meanwhile, we also propose an inner-dimensional evaluation strategy (INES) to efficiently evaluate the local importance of units inside each dimension with the gradient. By only performing ITES on the unit with the lowest local importance in each dimension, we are able to efficiently identify the redundant unit to prune and thus accelerate the pruning. 
% Based on INES and ITES, we further propose a heuristic pruning algorithm to progressively prune CNNs along the best trade-off between accuracy and model overhead, which iteratively utilizes INES and ITES to identify and remove redundant units from the three dimensions. By this means, our pruning framework efficiently explores the huge design space formed by the three dimensions and finds the optimal tiny model for edge devices. Extensive experiments validates the efficacy and efficiency of our approach.

In this paper, we present a multi-dimensional pruning framework, \mnade, to jointly prune the three dimensions of CNNs for embedded devices. First, we introduce an inter-dimensional evaluation strategy (ITES), which enables comprehensive evaluation of units across different dimensions with a novel metric named global importance, thereby accurately identifying the redundancy in the three dimensions. Meanwhile, we also propose an inner-dimensional evaluation strategy (INES) to efficiently evaluate units inside each dimension with local importance. By collaboratively using ITES and INES, we accurately and efficiently identify the redundancy in the three dimensions. Based on INES and ITES, we further propose a heuristic pruning algorithm, which utilizes INES and ITES to progressively identify and prune redundant units in the three dimensions. By this means, our pruning framework efficiently explores the huge design space formed by the three dimensions and finds the optimal tiny model for embedded devices. Extensive experiments validate the efficacy and efficiency of our approach.

\vspace{5pt}
\section*{Acknowledgement\vspace{5pt}}
This study is partially supported under the RIE2020 Industry Alignment Fund – Industry Collaboration Projects (IAF-ICP) Funding Initiative, as well as cash and in-kind contribution from the industry partner, HP Inc., through the HP-NTU Digital Manufacturing Corporate Lab (I1801E0028). This work is also partially supported by the Ministry of Education, Singapore, under its Academic Research Fund Tier 2 (MOE2019-T2-1-071), and Nanyang Technological University, Singapore, under its NAP (M4082282).
\vspace{5pt}

\bibliographystyle{IEEEtran}
\bibliography{reference}

\end{document}